\documentclass[10pt,twocolumn,letterpaper]{article}

\usepackage[pagenumbers]{iccv} 

\newcommand{\cparagraph}[1]{{\vspace{+1mm}\noindent\textbf{#1}\ }}

\newcommand{\cgreen}[1]{\textcolor{teal}{#1}}
\newcommand{\cred}[1]{\textcolor{Mahogany}{#1}}
\newcommand{\cblue}[1]{\textcolor{RoyalBlue}{#1}}

\definecolor{iccvblue}{rgb}{0.21,0.49,0.74}
\usepackage[pagebackref,breaklinks,colorlinks,allcolors=iccvblue]{hyperref}

\usepackage{bm}
\usepackage{pifont}
\usepackage{graphicx}
\usepackage{bbding}

\def\paperID{894} 
\def\confName{ICCV}
\def\confYear{2025}

\title{Personalize Anything for Free with Diffusion Transformer}

\vspace{-0.8em} 
\author{
\vspace{0.5em}Haoran Feng\textsuperscript{1}\footnotemark[1]~ \quad
Zehuan Huang\textsuperscript{2}\footnotemark[1]\footnotemark[2]~ \quad
Lin Li\textsuperscript{3}~ \quad 
Hairong Lv\textsuperscript{1}~ \quad 
Lu Sheng\textsuperscript{2}\ding{41}\\
\vspace{0.5em}\small $^{1}$Tsinghua University~~
$^{2}$Beihang University~~
\small $^{3}$Renmin University of China\\
\small Project page:~\url{https://fenghora.github.io/Personalize-Anything-Page/}
}
\vspace{-0.8em} 

\begin{document}

\twocolumn[
    \maketitle
    \vspace{-2.8em}
    \begin{center}
\centering
\includegraphics[width=\textwidth]{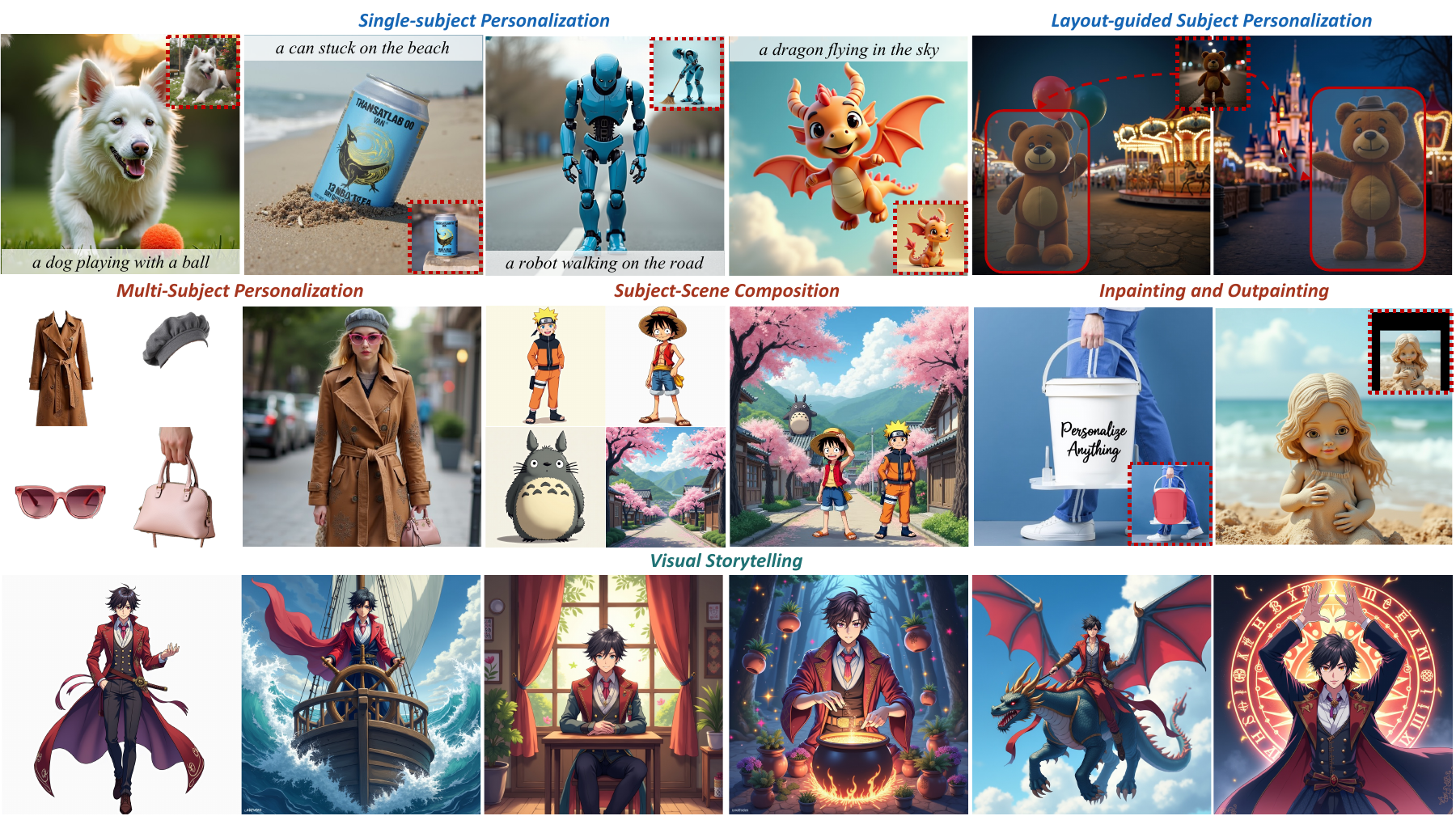}
\captionof{figure}{\textit{Personalize Anything} is a training-free framework based on Diffusion Transformers (DiT) for personalized image generation. The framework demonstrates advanced \textbf{versatility}, excelling in \cblue{\textit{\textbf{single-subject personalization} (top)}}, \cred{\textit{\textbf{multi-subject or subject-scene composition, inpainting and outpainting} (middle)}}, as well as applications like \cgreen{\textit{\textbf{visual storytelling} (bottom)}}, all without any training or fine-tuning.}
\label{fig:teaser}
\end{center}
    \bigbreak
]

\let\thefootnote\relax\footnotetext{
$^*$ Equal contribution \hspace{5pt}
$^\dagger$ Project lead \hspace{5pt} 
\ding{41} Corresponding author
}

\begin{abstract}
Personalized image generation aims to produce images of user-specified concepts while enabling flexible editing.
Recent training-free approaches, while exhibit higher computational efficiency than training-based methods, struggle with identity preservation, applicability, and compatibility with diffusion transformers (DiTs).
In this paper, we uncover the untapped potential of DiT, where simply replacing denoising tokens with those of a reference subject achieves zero-shot subject reconstruction.
This simple yet effective feature injection technique unlocks diverse scenarios, from personalization to image editing.
Building upon this observation, we propose \textbf{Personalize Anything}, a training-free framework that achieves personalized image generation in DiT through:
1) timestep-adaptive token replacement that enforces subject consistency via early-stage injection and enhances flexibility through late-stage regularization, 
and 2) patch perturbation strategies to boost structural diversity. 
Our method seamlessly supports layout-guided generation, multi-subject personalization, and mask-controlled editing.
Evaluations demonstrate state-of-the-art performance in identity preservation and versatility.
Our work establishes new insights into DiTs while delivering a practical paradigm for efficient personalization.
\end{abstract}

\section{Introduction}
\label{sec:intro}

Personalized image generation aims to synthesize images of user-specified concepts while enabling flexible editing.
The advent of text-to-image diffusion models~\cite{ramesh2022dalle2,nichol2022glide,saharia2022imagen,dhariwal2021diffusionbeatgans,podell2023sdxl,peebles2023dit,flux,kolors,ldm} has revolutionized this field, enabling applications in areas like advertising production.

Previous research on subject image personalization relies on test-time optimization or large-scale fine-tuning.
Optimization-based approach~\cite{Dreambooth,MultiConcept,MixOfShow,Cones2,ConceptWeaver,Omg,multibooth,Ziplora,Magicapture} enables the pre-trained models to learn the specific concept through fine-tuning on a few images of a subject.
While achieving identity preservation, these methods demand substantial computational resources and time due to per-subject optimization requiring hundreds of iterative steps.
Large-scale fine-tuning alternatives~\cite{BlipDiffusion,IpAdapter,SubjectDiffusion,SsrEncoder,lambdaECLIPSE,mc2,Moma,CSGO,msdiffusion,pOps,EZIGen,mipadapter,Ominicontrol,OneDiffusion,UniReal,Easyref,AnyStory,ObjectLevel,NestedAttention,Photomaker,Instantid,FlashFace,Consistentid,Parts2whole,Pulid,LcmLookahead,CharacterFactory,chen2024anydoor} seek to circumvent this limitation by training auxiliary networks on large-scale datasets to encode reference images.
However, these approaches both demand heavy training requirements and risk overfitting to narrow data distributions, degrading their generalizability.

Recent training-free solutions~\cite{consistory,StoryDiffusion,BC2NewStories,ChosenOne,FreeCustom,LargeScale,TweedieMix} exhibit higher computational efficiency than the training-based approach.
These methods typically leverage an attention sharing mechanism to inject reference features, processing denoising and reference subject tokens jointly in pre-trained self-attention layers.
However, these attention-based methods lack constraints on subject consistency and often fail to preserve identity.
Moreover, their application to advanced text-to-image diffusion transformers (DiTs)~\cite{peebles2023dit,flux,sd3} proves challenging stemming from DiT's positional encoding mechanism.
As analyzed in \cref{subsec:exploration}, we attribute this limitation to the strong influence of the explicitly encoded positional information on DiT's attention mechanism.
This makes it difficult for generated images to correctly attend to the reference subject's tokens within traditional attention sharing.

\begin{figure}
    \centering
    \includegraphics[width=\linewidth]{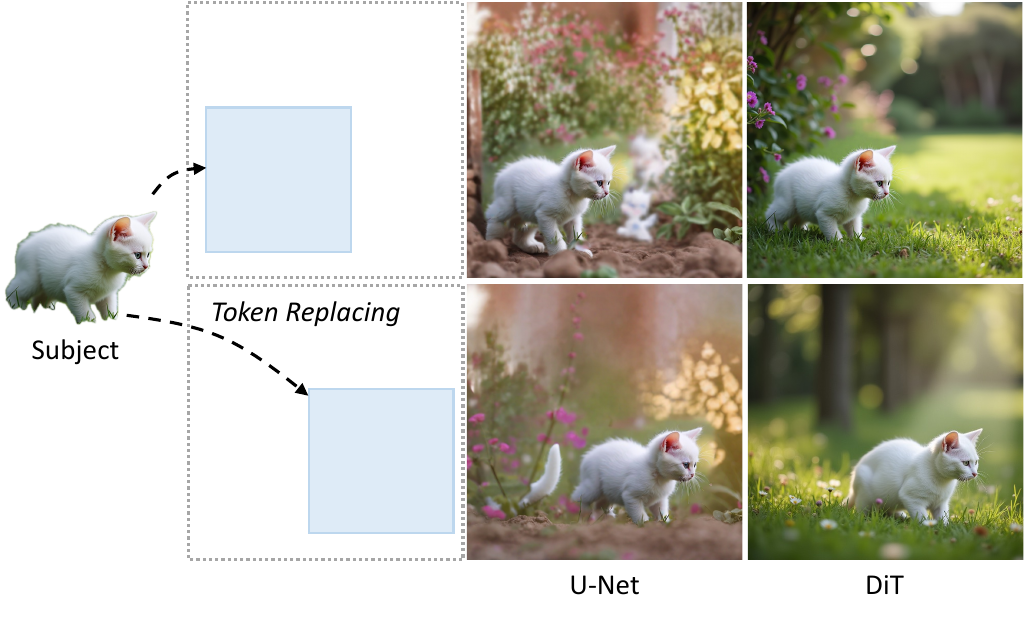}
    \caption{Simple token replacement in DiT (right) achieves high-fidelity subject reconstruction through its position-disentangled representation, while U-Net's convolutional entanglement (left) induces blurred edges and artifacts.}
    \label{fig:finding_replace}
\end{figure}

In this paper, we delve into the diffusion transformers (DiTs)~\cite{peebles2023dit,flux,sd3}, and observe that simply replacing the denoising tokens with those of a reference subject allows for high-fidelity subject reconstruction.
As illustrated in \cref{fig:finding_replace}, DiT exhibits exceptional reconstruction fidelity under this manipulation, while U-Net~\cite{unet} often induces blurred edges and artifacts.
We attribute this to the separate embedding of positional information in DiT, achieved via its explicit positional encoding mechanism.
This decoupling of semantic features and position enables the substitution of purely semantic tokens, avoiding positional interference.
Conversely, U-Net's convolutional mechanism binds texture and spatial position together, causing positional conflicts when replacing tokens and leading to low-quality image generation.
This discovery establishes token replacement as a viable pathway for zero-shot subject personalization in DiT, unlocking various scenarios ranging from personalization to inpainting and outpainting, without necessitating complicated attention engineering.

Building on this foundation, we propose ``\textbf{Personalize Anything}'', a training-free framework enabling personalized image generation in DiT through timestep-adaptive token replacement and patch perturbation strategies.
Specifically, we inject reference subject tokens (excluding positional information) in the earlier steps of the denoising process to enforce subject consistency, while enhancing flexibility in the later steps through multi-modal attention.
Furthermore, we introduce patch perturbation to the reference tokens before token replacement, locally shuffling them and applying morphological operations to the subject mask.
It encourages the model to introduce more global appearance information and enhances structural and textural diversity.
Additionally, our framework seamlessly supports
1) layout-guided generation through translations on replacing regions,
2) multi-subject personalization and subject-scene composition via sequential injection of reference subjects or scene,
and 3) extended applications (\eg inpainting and outpainting) via incorporating user-specified mask conditions.

Comprehensive evaluations on multiple personalization tasks demonstrate that our training-free method exhibits superior identity preservation, fidelity and versatility, outperforming existing approaches including those fine-tuned on DiTs.
Our contributions are summarized as follows:
\begin{itemize}
    \item We uncover DiT's potential for high-fidelity subject reconstruction via simple token replacement, and characterize its position-disentangled properties.
    \item We introduce a simple yet effective framework, denoted as ``Personalize Anything'', which starts with subject reconstruction and enhances the flexibility via timestep-adaptive replacement and patch perturbation.
    \item Experiments demonstrates that the proposed framework exhibits high consistency, fidelity, and versatility across multiple personalization tasks and applications.
\end{itemize}

\section{Related Work}
\label{sec:related}

\subsection{Text-to-Image Diffusion Models}

Text-to-image generation has been revolutionized by diffusion models~\cite{ddpm,ddim} that progressively denoise Gaussian distributions into images.
Among a series of effective works~\cite{{ramesh2022dalle2,nichol2022glide,saharia2022imagen,dhariwal2021diffusionbeatgans,podell2023sdxl,peebles2023dit,flux,kolors,ldm}}, Latent Diffusion Model~\cite{ldm} employs a U-Net backbone~\cite{unet} 
for efficient denoising within compressed latent space, becoming the foundation for subsequent improvements in resolution~\cite{podell2023sdxl}.
A recent architectural shift replaces convolutional U-Nets with vision transformers~\cite{dosovitskiy2020vit,peebles2023dit}, exploiting their global attention mechanisms and geometrically-aware positional encodings.
These diffusion transformers (DiTs) demonstrate superior scalability~\cite{sd3,flux}—performance improvements consistently correlate with increased model capacity and training compute, establishing them as the new state-of-the-art paradigm. 

\subsection{Personalized Image Generation}

\cparagraph{Training-Based Approaches.}
Previous subject personalization methods primarily adopt two strategies:
i) Test-time optimization techniques~\cite{Dreambooth,oneimgoneword,MultiConcept,MixOfShow,Cones2,ConceptWeaver,Omg,multibooth,Ziplora,Magicapture,pplus,svdiff,keylocked} that fine-tune foundation models on target concepts at inference time, often requiring 30 GPU-minute optimization per subject;
and ii) large-scale training-based methods~\cite{BlipDiffusion,IpAdapter,SubjectDiffusion,SsrEncoder,lambdaECLIPSE,mc2,Moma,CSGO,msdiffusion,pOps,EZIGen,mipadapter,Ominicontrol,OneDiffusion,UniReal,Easyref,AnyStory,ObjectLevel,NestedAttention,Photomaker,Instantid,FlashFace,Consistentid,Parts2whole,Pulid,LcmLookahead,CharacterFactory,chen2024anydoor,msdiffusion} that learn concept embeddings through auxiliary networks pre-trained on large datasets.
While achieving notable fidelity, both paradigms suffer from computational overheads and distribution shifts that limit real-world application.

\cparagraph{Training-Free Alternatives.}
Emerging training-free methods~\cite{consistory,StoryDiffusion,BC2NewStories,ChosenOne,FreeCustom,LargeScale,TweedieMix} exhibit higher computational efficiency than the training-based approach.
These methods typically leverage an attention sharing mechanism to inject reference features, processing denoising and reference subject tokens jointly in pre-trained self-attention layers.
However, these attention-based methods lack constraints on subject consistency and fail to preserve identity.
Moreover, their application to advanced diffusion transformers (DiTs)~\cite{peebles2023dit,flux,sd3} proves challenging due to DiT's explicit positional encoding, thereby limiting their scalability to larger-scale text-to-image generation models~\cite{flux,sd3}.


\section{Methodology}
\label{sec:method}

This paper introduces a training-free paradigm for personalized generation using diffusion transformers (DiTs)~\cite{peebles2023dit}, synthesizing high-fidelity depictions of user-specified concepts while preserving textual controllability.
In the following sections, we start with an overview of 
standard architectures in text-to-image diffusion models.
\cref{subsec:exploration} systematically reveals architectural distinctions that impede the application of existing attention sharing mechanisms to DiTs.
\cref{subsec:analysis} uncovers DiT's potential for subject reconstruction via simple token replacement, culminating the presentation of our \textit{Personalize Anything} framework in \cref{subsec:framework}.

\subsection{Preliminaries}
\label{subsec:pre}

Diffusion models progressively denoise a latent variable $z_{T}$ through a network $\epsilon_{\theta}$, with architectural choices being of paramount importance. We analyze two main paradigms:

\cparagraph{U-Net Architectures.}
The convolutional U-Net~\cite{unet} in Stable Diffusion~\cite{ldm} comprises pairs of down-sampling and up-sampling blocks connected by a middle block.
Each block interleaves residual blocks for feature extraction with spatial attention layers for capturing spatial relationships.

\cparagraph{Diffusion Transformer (DiT).}
Modern DiTs~\cite{peebles2023dit} in advanced models~\cite{flux,sd3} leverage transformers~\cite{dosovitskiy2020vit} to process discretized latent representations, including image tokens $X\in\mathbb{R}^{N\times d}$ and text tokens $C\in\mathbb{R}^{M\times d}$, where $d$ is the embedding dimension, $N$ and $M$ are the length of sequences.
These models typically encode positional information of $X$ through RoPE~\cite{su2024rope}, which applies rotation matrices based on the token's coordinate $(i,j)$ in the 2D grid:
\begin{equation}
    \dot{X}^{i,j} = X^{i,j} \cdot \bm{R}(i,j)
\end{equation}
where $\bm{R}(i,j)$ denotes the rotation matrix at position $(i,j)$ with $0\leq i < w$ and $0\leq j<h$.
Text tokens $C$ receive fixed positional anchors $(i=0,j=0)$ to maintain modality distinction.
The multi-modal attention mechanism (MMA) is then applied to all position-encoded tokens $[\dot{X};\dot{C}_T]\in\mathbb{R}^{(N+M)\times d}$, enabling full bidirectional attention across both modalities.

\subsection{Attention Sharing Fails in DiT}
\label{subsec:exploration}

We systematically investigate why established U-Net-based personalization techniques~\cite{FreeCustom,consistory} fail when naively applied to DiT architectures~\cite{flux}, identifying positional encoding conflicts as the core challenge.

\begin{figure}
    \centering
    \includegraphics[width=\linewidth]{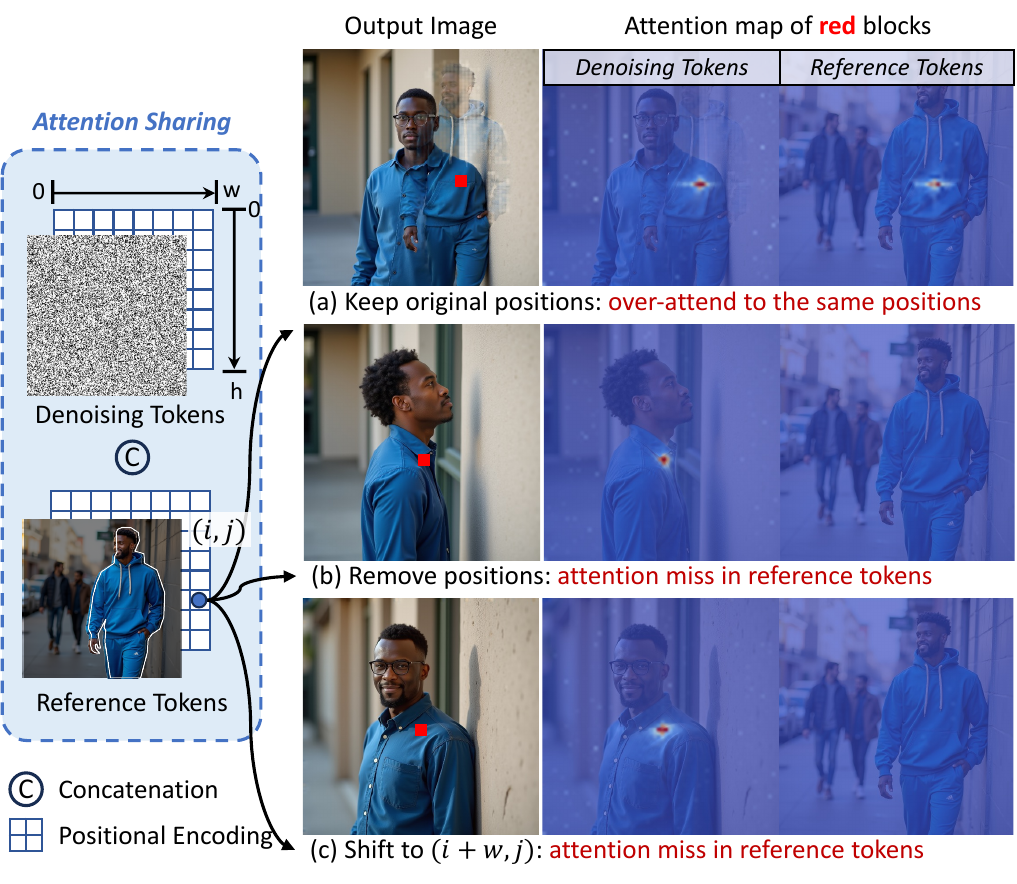}
    \caption{Attention sharing~\cite{consistory,FreeCustom} fails in DiT due to the explicit positional encoding mechanism. When keeping the original positions $(i,j)\in [0,w)\times[0,h)$ in reference tokens, denoising tokens over-attend to reference ones with the same positions (shown in attention maps of (a)), resulting in ghosting artifacts in the generated image. Modified strategies, (b) removing positions and (c) shifting to non-overlapping regions, avoid collisions but loses identity alignment, as attention is almost absent on reference tokens.}
    \label{fig:finding_positions}
\end{figure}

\cparagraph{Positional Encoding Collision.}
Implementing attention sharing of existing methods~\cite{consistory,FreeCustom} in DiT, we concatenate position-encoded denoising tokens $\dot{X}$ and reference tokens $\dot{X}_{ref}$ (obtained via flow inversion~\cite{TamingRectifiedFlow}) into a unified sequence $[\dot{X};\dot{X}_{ref}]$.
Both tokens keep the original positions $(i,j)\in [0,w)\times[0,h)$, causing destructive interference to attention computation.
As visualized in \cref{fig:finding_positions}a, this forces denoising tokens to over-attend to reference tokens with the same positions, resulting in ghosting artifacts of the reference subject in the generated image.
Quantitative analysis in supplementary materials reveals that the attention score between denoising and reference tokens at the same position in DiT is \textbf{723\%} higher than in U-Net, confirming DiT's position sensitivity.

\cparagraph{Modified Encoding Strategies.}
Motivated by DiT's position-disentangled encoding (\cref{subsec:pre}), we engineer two positional adjustments on $X_{ref}$ to obtain non-conflicting $\dot{X}_{ref}$:
i) remove positions and fix all reference positions to $(0,0)$ akin to text tokens,
and ii) shift reference tokens to $(i',j')=(i+w,j)$, creating non-overlapping regions.
As shown in \cref{fig:finding_positions} (b) and (c), while eliminating collisions, both methods struggle to preserve identity, as attention is almost absent on reference tokens.

In summary, the explicitly encoded positional information exhibits strong influence on the attention mechanism in DiT—a fundamental divergence from U-Net's implicit position handling.
This makes it difficult for generated images to correctly attend to the reference subject's tokens within traditional attention sharing.

\begin{figure*}
    \centering
    \includegraphics[width=\textwidth]{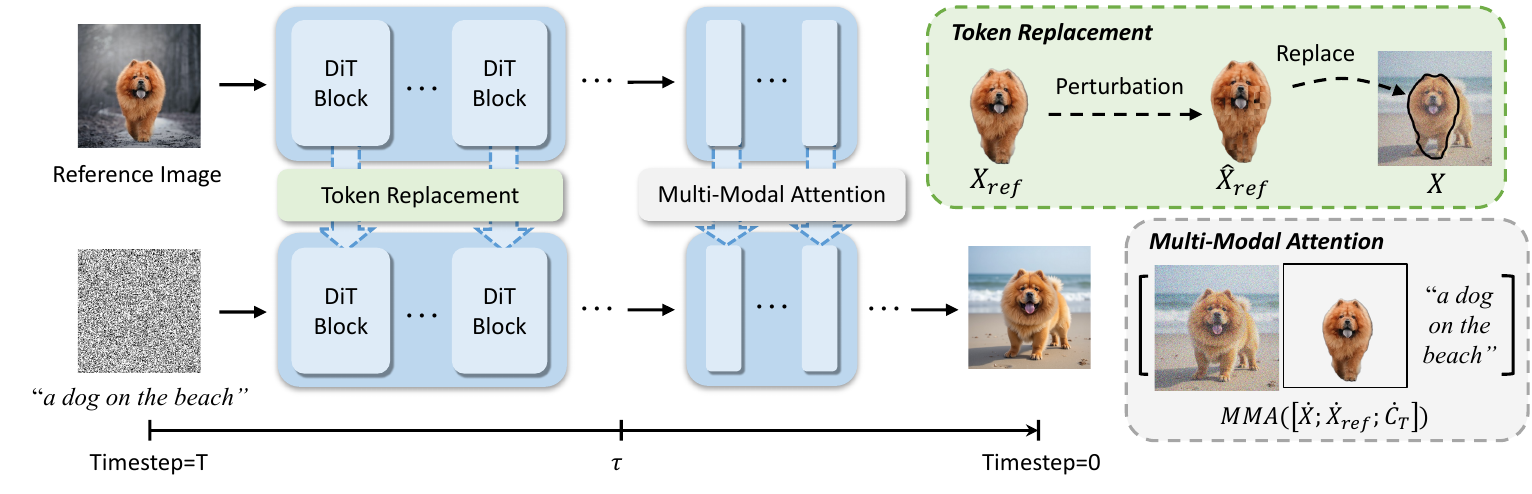}
    \caption{Method overview. Our framework anchors subject identity in early denoising through mask-guided token replacement with preserved positional encoding, and transitions to multi-modal attention for semantic fusion with text in later steps. During token replacement, we inject variations via patch perturbations. This timestep-adaptive strategy balances identity preservation and generative flexibility.}
    \label{fig:overview}
\end{figure*}

\begin{figure*}
    \centering
    \includegraphics[width=\textwidth]{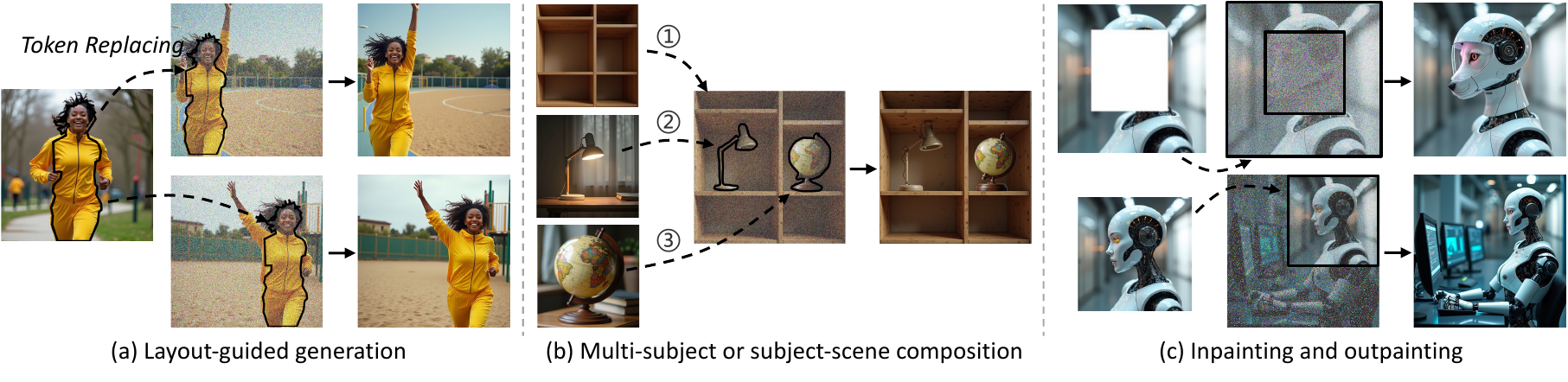}
    \caption{Seamless extensions. Our framework enables: (a) layout-guided generation by translating token-injected regions, (b) multi-subject composition through sequential token injection, and (c) inpainting and outpainting via specifying masks and increased replacement.}
    \label{fig:method_extensions}
\end{figure*}

\subsection{Token Replacement in DiT}
\label{subsec:analysis}

Building on the foundational observation on DiT's architectural distinctions, we extend our investigation to the latent representation in DiT.
We uncover that simply replacing the denoising tokens with those of a reference subject allows for high-fidelity subject reconstruction.

Specifically, we apply inversion techniques~\cite{TamingRectifiedFlow,SemanticImageInversion} on the reference image, obtaining the reference tokens $X_{ref}$ without encoded positional information, as well as the reference subject's mask $\mathcal{M}_{ref}$.
We then inject $X_{ref}$ into specific region $\mathcal{M}$ of the denoising tokens $X$ via token replacement:
\begin{equation}
\label{eq:replace}
\hat{X} = X \odot (1-\mathcal{M}) + X_{ref} \odot \mathcal{M}
\end{equation}
where $\mathcal{M}$ can be obtained by translating $\mathcal{M}_{ref}$.
As shown in \cref{fig:finding_replace}, token replacement in DiT reconstructs high-fidelity images with consistent subjects in specified positions, while U-Net's convolutional entanglement manifests as blurred edges and artifacts.

We attribute this to the separate embedding of positional information in DiT, achieved via its explicit positional encoding mechanism (\cref{subsec:pre}).
This decoupling of semantic features and position enables the substitution of purely semantic tokens, avoiding positional interference.
Conversely, U-Net's convolutional mechanism binds texture and spatial position together, causing positional conflicts when replacing tokens and leading to low-quality image generation.
This discovery establishes token replacement as a viable pathway for zero-shot subject personalization in DiT.
It unlocks various scenarios ranging from personalization to inpainting and outpainting, without necessitating complicated attention engineering, and establishes the foundation for our personalization framework in \cref{subsec:framework}.






\subsection{Personalize Anything}
\label{subsec:framework}
Building upon these discoveries, we propose \textit{Personalize Anything}, a novel training-free personalization for diffusion transformers (DiTs).
This framework draws inspiration from zero-shot subject reconstruction in DiTs, and effectively enhances flexibility by timestep-adaptive token replacement and patch perturbation strategies (\cref{fig:overview}).

\cparagraph{Timestep-adaptive Token Replacement.}
Our method begins by inverting a reference image containing the desired subject~\cite{TamingRectifiedFlow}.
This process yields reference tokens $X_{ref}$ (excluding positional encodings) and a corresponding subject mask $\mathcal{M}_{ref}$~\cite{DiffusionIsSegmenter}.
Instead of continuous injection throughout the denoising process as employed in subject reconstruction, we introduce a timestep-dependent strategy:

\ding{182} \textbf{\emph{Early-stage subject anchoring via token replacement ($t>\tau$).}}
During the initial denoising steps ($t>\tau$, where $\tau$ is an empirically determined threshold set at 80\% of the total denoising steps $T$), we anchor the subject's identity by replacing the denoising tokens $X$ within the subject region $\mathcal{M}$ with the reference tokens $X_{ref}$ (\cref{eq:replace}).
The region $\mathcal{M}$ can be obtained by translating $\mathcal{M}_{ref}$ to the user-specified location.
We preserve the positional encodings associated with the denoising tokens $X$ to maintain spatial coherence.

\ding{183} \textbf{\emph{Later-stage semantic fusion via multi-modal attention ($t\leq\tau$).}}
In later denoising steps $t\leq\tau$, we transition to semantic fusion.
Here we concatenate zero-positioned reference tokens $\dot{X}_{ref}$ with denoising tokens $\dot{X}$ and text embeddings $\dot{C}$.
The unified sequence $[\dot{X};\dot{X}_{ref};\dot{C}_{T}]$ undergoes Multi-Modal Attention (MMA) to harmonize subject guidance with textual conditions.
This adaptive threshold $\tau$ balances the preservation of subject identity with the flexibility afforded by the text prompt.

\cparagraph{Patch Perturbation for Variation.}
To prevent identity overfitting while preserving identity consistency, we introduce two complementary perturbations:
1) Random Local token shuffling within 3x3 windows disrupts rigid texture alignment,
and 2) Mask augmentation of $\mathcal{M}_{ref}$, including simulating natural shape variations using morphological dilation/erosion with a 5px kernel, or manual selection of regions emphasizing identity.
The idea behind this local interference technique is to encourage the model to introduce more global textural features while enhancing structural and local diversity.

\cparagraph{Seamless Extensions.}
As illustrated in ~\cref{fig:method_extensions}, our framework naturally extends to complex scenarios through geometric programming:
Translating $\mathcal{M}$ enables the spatial arrangement of subjects thereby achieving layout-guided generation,
while sequential injection of multiple $\{ X_{ref}^{k} \}$ into disjoint $\{ \mathcal{M}_{k} \}$ regions and unified Multi-Modal Attention $\text{MMA}([\dot{X};\{ \dot{X}_{ref}^{k} \};\dot{C}_{T}])$ facilitate multi-subject or subject-scene composition.
For image editing tasks, we incorporate user-specified masks in the inversion process to obtain reference $X_{ref}$ and $\mathcal{M}_{ref}$ that should be preserved.
Meanwhile, we disable perturbations and set $\tau$ to 10\% total steps, preserving the original image content as much as possible and achieving coherent inpainting or outpainting.

\section{Experiments}
\label{sec:exp}

\begin{figure*}
    \centering
    \includegraphics[width=\textwidth]{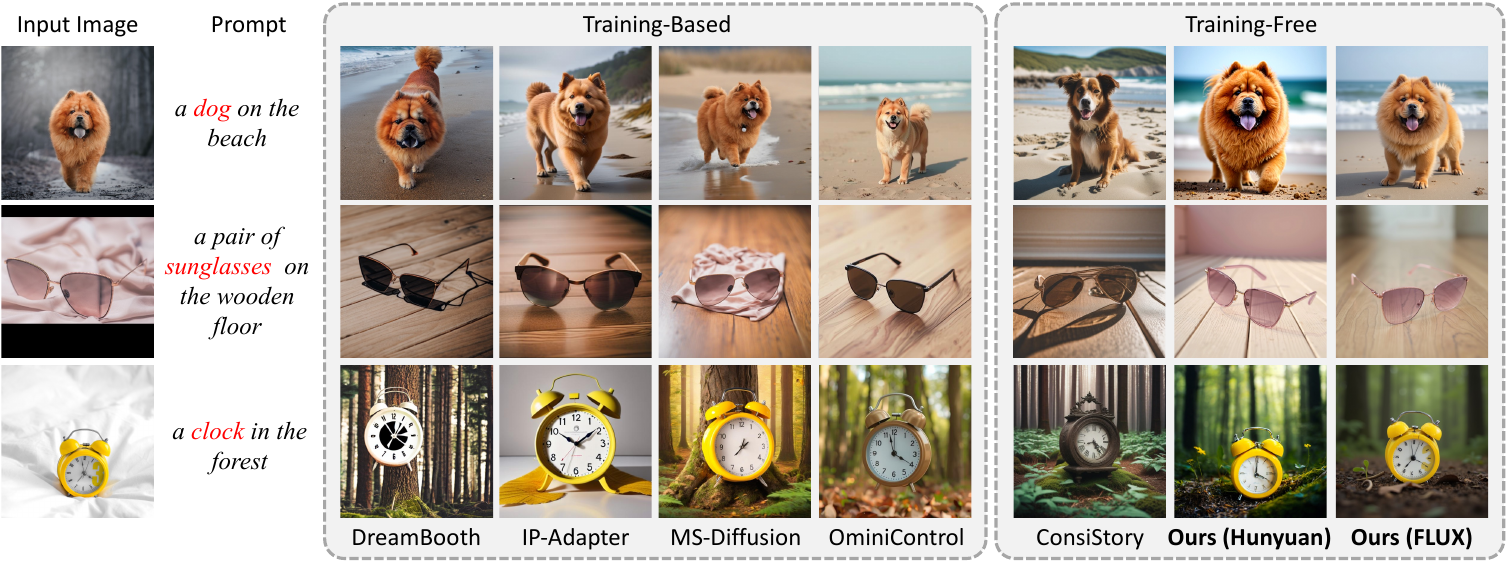}
    \caption{Qualitative comparisons on single-subject personalization. More results can be found in the supplementary materials.}
    \label{fig:comparison_single_subject}
\end{figure*}

\subsection{Experimental Setup}

\cparagraph{Implementation Details.}
Our framework builds upon the open-source HunyuanDiT~\cite{hunyuandit} and FLUX.1-dev~\cite{flux}.
We adopt 50-step sampling with classifier-free guidance ($w=3.5$), generating $1024\times1024$ resolution images.
Token replacement threshold $\tau$ is set to 80\% total steps.

\cparagraph{Benchmark Protocols.}
We establish three evaluation tiers:
1) Single-subject personalization, compared against 10 approaches spanning training-based~\cite{msdiffusion,Ominicontrol,OneDiffusion,EZIGen,lambdaECLIPSE,SsrEncoder,BlipDiffusion,IpAdapter,Dreambooth} and training-free~\cite{consistory} paradigms,
2) Multi-subject personalization, evaluated against 6 representative methods~\cite{msdiffusion,lambdaECLIPSE,MultiConcept,Cones2,mipadapter,FreeCustom},
and 3) Subject-scene composition, benchmarked using AnyDoor~\cite{chen2024anydoor} as reference for contextual adaptation.

\cparagraph{Evaluation Metrics.}
We evaluate our \textit{Personalize Anything} on DreamBench~\cite{Dreambooth} which comprises $30$ base objects each accompanied by $25$ textual prompts.
We extend this dataset to 750, 1000, and 100 test cases for single-subject, multi-subject, and subject-scene personalization using combinatorial rules.
Quantitative assessment leverages multi-dimensional metrics:
FID~\cite{fid} for quality analysis,
CLIP-T~\cite{clip} for image-text alignment,
and DINO~\cite{dinov2}, CLIP-I~\cite{clip}, DreamSim~\cite{dreamsim} for identity preservation in single-subject evaluation while SegCLIP-I~\cite{multibooth} in multi-subject evaluation.
DreamSim~\cite{dreamsim} is a new metric for perceptual image similarity that bridges the gap between ``low-level'' metrics (e.g., PSNR, SSIM, LPIPS~\cite{lpips}) and ``high-level'' measures (\eg CLIP~\cite{clip}).
SegCLIP-I is similar to CLIP-I, but all the subjects in source images are segmented.

\subsection{Comparison to State-of-the-Arts}

\cparagraph{Single-Subject Personalization.}
\cref{fig:comparison_single_subject} shows qualitative comparison with representative baseline methods.
Existing test-time fine-tuning methods~\cite{Dreambooth} require 30 GPU-minute optimization for each concept and sometimes exhibit concept confusion for single-image inputs, manifesting as treating the background color as a characteristic of the subject.
Training-based but test-time tuning-free methods~\cite{IpAdapter,OneDiffusion,Ominicontrol,msdiffusion}, despite trained on large datasets, struggle to preserve identity in detail for real image inputs.
Training-free methods~\cite{consistory} generate inconsistent subjects with single-image input.
In contrast, our method produces high-fidelity images that are highly consistent with the specified subjects, without necessitating training or fine-tuning.
Quantitative results in \cref{tab:comparison_single_subject} confirms our excellent performance on identity preservation and image-text alignment.   

\begin{figure*}
    \centering
    \includegraphics[width=\textwidth]{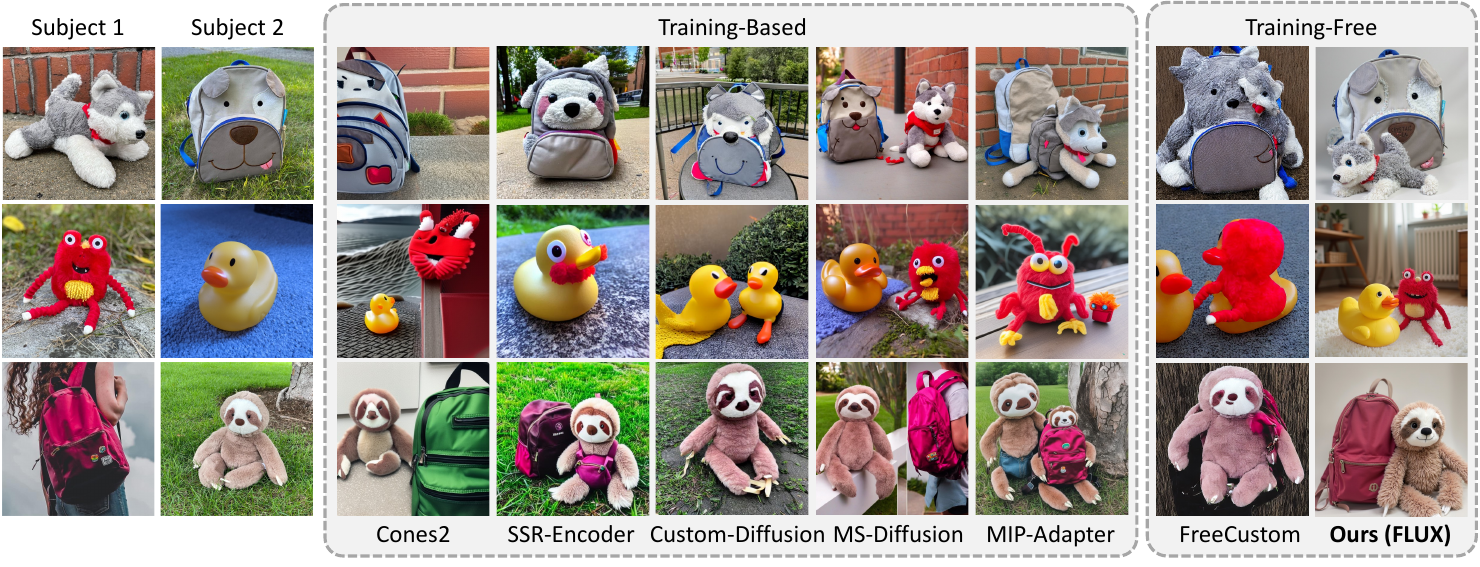}
    \caption{Qualitative comparisons on multi-subject personalization.}
    \label{fig:comparison_multi_subject}
\end{figure*}

\begin{figure}
    \centering
    \includegraphics[width=\linewidth]{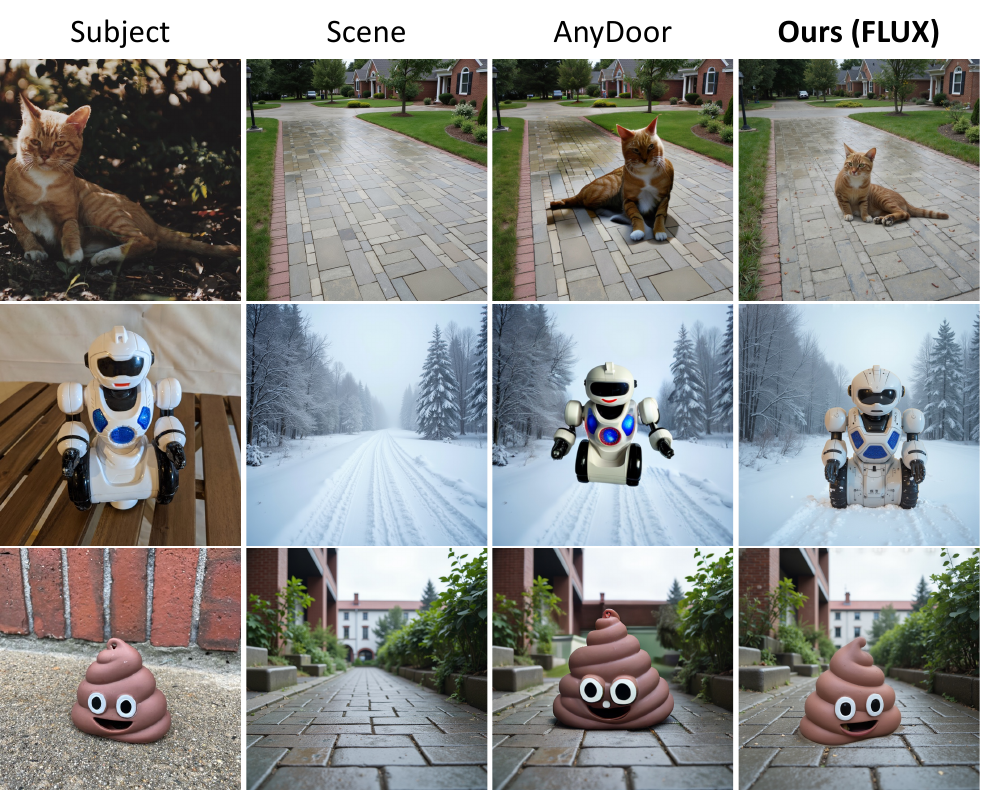}
    \caption{Qualitative results on subject-scene composition.}
    \label{fig:comparison_subject_scene}
\end{figure}

\begin{table}\footnotesize
    \caption{Quantitative results on single-subject personalization.}
    \label{tab:comparison_single_subject}
        \centering
        \setlength{\tabcolsep}{3pt}
        \begin{tabular}{l|cccc}
        \toprule
        Method & CLIP-T$\uparrow$ & CLIP-I$\uparrow$ & DINO$\uparrow$ & DreamSim$\downarrow$  \\
        \midrule
        DreamBooth~\cite{Dreambooth} & 0.271 & 0.819 & 0.550 & 0.290  \\
        BLIP-Diffusion~\cite{BlipDiffusion} & 0.251 & 0.835 & 0.641 & 0.283  \\
        IP-Adapter~\cite{IpAdapter} & 0.249 & 0.861 & 0.652 & 0.256  \\
        $\lambda$-ECLIPSE~\cite{lambdaECLIPSE} & 0.235 & 0.866 & 0.682 & 0.224  \\
        SSR-Encoder~\cite{SsrEncoder} & 0.244 & 0.860 & \textbf{0.701} & 0.220  \\
        EZIGen~\cite{EZIGen} & 0.263 & 0.825 & 0.662 & 0.247  \\
        MS-Diffusion~\cite{msdiffusion} & 0.283 & 0.824 & 0.539 & 0.261  \\
        OneDiffusion~\cite{OneDiffusion} & 0.255 & 0.817 & 0.603 & 0.298  \\
        OminiControl~\cite{Ominicontrol} & 0.275 & 0.820 & 0.516 & 0.301  \\
        \midrule
        ConsiStory~\cite{consistory} & 0.284 & 0.753 & 0.472 & 0.434  \\
        \textbf{Ours (HunyuanDiT~\cite{hunyuandit})} & 0.291 & 0.869 & 0.679 & 0.206 \\
        \textbf{Ours (FLUX~\cite{flux})} & \textbf{0.307} & \textbf{0.876} & 0.683 & \textbf{0.179}  \\
       \bottomrule
    \end{tabular}
\end{table}

\cparagraph{Multi-Subject Personalization.}
From a qualitative perspective in \cref{fig:comparison_multi_subject}, existing approaches~\cite{Cones2,SsrEncoder,MultiConcept,msdiffusion,mipadapter} may suffer from conceptual fusion when generating multiple subjects, struggling to maintain their individual identities, or produce fragmented results due to incorrect modeling of inter-subject relationships.
In contrast, our approach manages to maintain natural interactions among subjects via layout-guided generation, while ensuring each subject retains its identical characteristics and distinctiveness.
Quantitatively, results in \cref{tab:comparison_multi_subject} demonstrate the strength of \textit{Personalize Anything} in SegCLIP-I and CLIP-T, demonstrating that our approach not only effectively captures identity of multiple subjects but also excellently preserves the text control capabilities.

\cparagraph{Subject-Scene Composition.}
We further evaluate our \textit{Personalize Anything} on subject-scene composition, conducting comparison with Anydoor~\cite{chen2024anydoor}.
We show the visualization results in \cref{fig:comparison_subject_scene}, where AnyDoor produces incoherent results, manifested as inconsistencies between the subject and environmental factors such as lighting in the generated images. 
In contrast, our method successfully generates natural images while effectively preserving the details of the subjects.
It demonstrates the huge potentials and generalization capabilities of \textit{Personalize Anything} in generating high-fidelity personalized images.

\begin{table}\small
    \caption{Quantitative results on multi-subject personalization.}
    \label{tab:comparison_multi_subject}
        \centering
        \setlength{\tabcolsep}{3pt}
        \begin{tabular}{l|ccc}
        \toprule
        Method & CLIP-T$\uparrow$ & CLIP-I$\uparrow$ & SegCLIP-I$\uparrow$ \\
        \midrule
        $\lambda$-ECLIPSE~\cite{lambdaECLIPSE} & 0.258 & 0.738 & 0.757  \\
        SSR-Encoder~\cite{SsrEncoder} & 0.234 & 0.720 & 0.761 \\
        Cones2~\cite{Cones2} & 0.255 & 0.747 & 0.702 \\
        Custom Diffusion~\cite{MultiConcept} & 0.228 & 0.727 & 0.781 \\
        MIP-Adapter~\cite{mipadapter} & 0.276 & 0.765 & 0.751 \\
        MS-Diffusion~\cite{msdiffusion} & 0.278 & 0.780 & 0.748 \\
        \midrule
        FreeCustom~\cite{FreeCustom} & 0.248 & 0.749 & 0.734 \\
        \textbf{Ours (HunyuanDiT)~\cite{hunyuandit}} & 0.284 & 0.817 & 0.832 \\
        \textbf{Ours (FLUX)} & \textbf{0.302} & \textbf{0.843} & \textbf{0.891}  \\
       \bottomrule
    \end{tabular}
\end{table}

\begin{table}\small
    \caption{Ablation studies. We evaluate the effects of token replacement threshold $\tau$ and patch perturbation.}
    \label{tab:ablation_study}
        \centering
        \setlength{\tabcolsep}{3pt}
        \begin{tabular}{lc|cccc}
        \toprule
        $\tau$ & Pertur. & CLIP-T$\uparrow$ & CLIP-I$\uparrow$ & DINO$\uparrow$ & DreamSim$\downarrow$ \\
        \midrule
        $T$ & \XSolidBrush & \textbf{0.317} & 0.764 & 0.625 & 0.305 \\
        $0.95\ T$ & \XSolidBrush & 0.313 & 0.773 & 0.632 & 0.294 \\
        $0.90\ T$ & \XSolidBrush & 0.306 & 0.849 & 0.680 & 0.199 \\
        $0.80\ T$ & \XSolidBrush & 0.302 & 0.882 & 0.741 & 0.163 \\
        $0.70\ T$ & \XSolidBrush & 0.282 & \textbf{0.920} & \textbf{0.769} & \textbf{0.140} \\
        \midrule
       $0.80\ T$ & \Checkmark & 0.307 & 0.876 & 0.683 & 0.179 \\
       \bottomrule
    \end{tabular}
\end{table}

\begin{figure*}
    \centering
    \includegraphics[width=\textwidth]{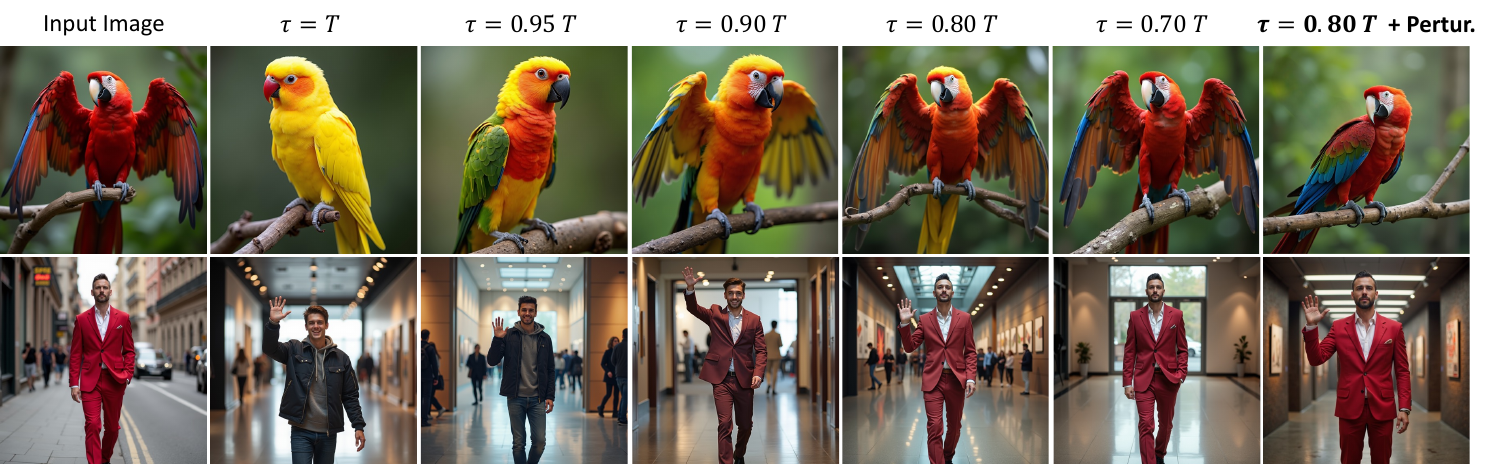}
    \caption{Qualitative ablation studies on token replacement threshold $\tau$ and patch perturbation.}
    \label{fig:ablation_study}
\end{figure*}

\begin{figure*}
    \centering
    \includegraphics[width=\textwidth]{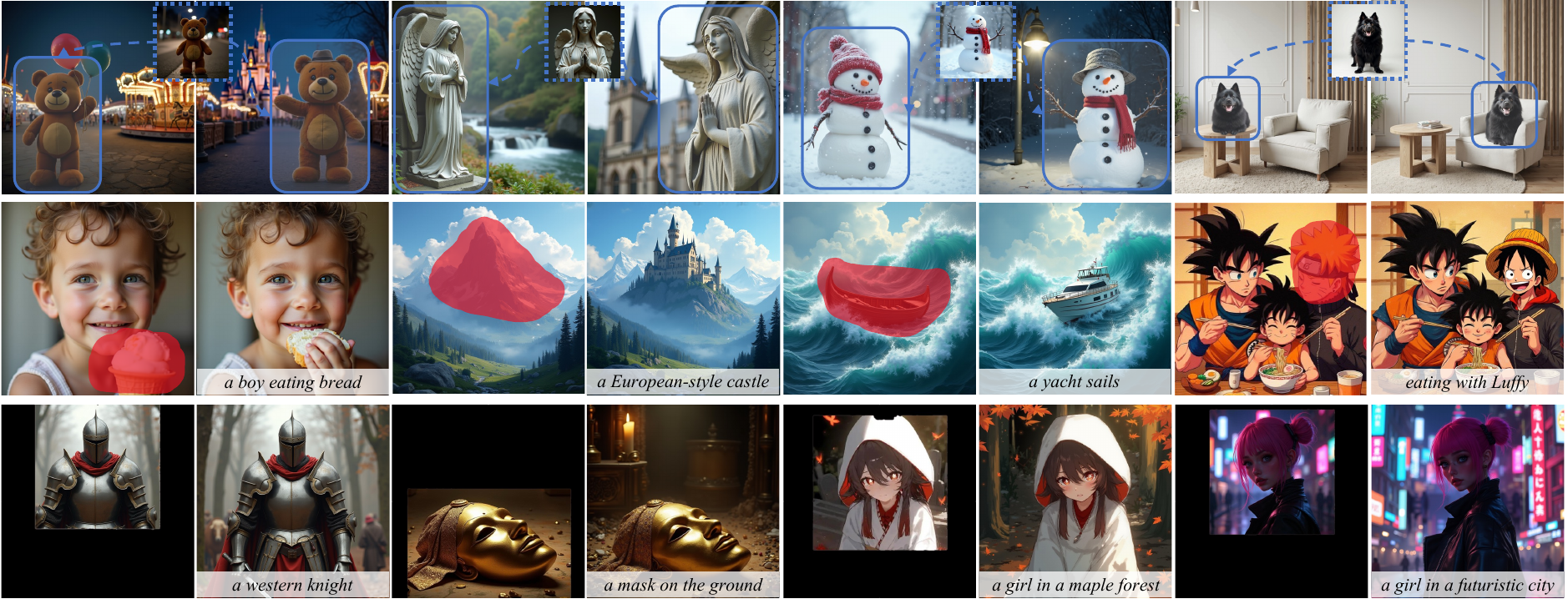}
    \caption{Visualization results of \textit{Personalize Anything} in \cblue{\textit{\textbf{layout-guided generation} (top)}}, \cred{\textit{\textbf{inpainting} (middle)}}, and \cgreen{\textit{\textbf{outpainting} (bottom)}}.}
    \label{fig:applications}
\end{figure*}

\subsection{Ablation Study}

We conduct ablation studies on single-subject personalization, examining the effects of token replacement timestep threshold $\tau$ and the patch perturbation strategy.

\cparagraph{Effects of Threshold $\tau$.}
Our systematic investigation of the timestep threshold $\tau$ reveals its critical role in balancing reference subject consistency and flexibility.
As visualized in \cref{fig:ablation_study}, early-to-mid replacement phases ($\tau>0.8\ T$) progressively incorporate geometric and appearance priors from reference tokens, initially capturing coarse layouts ($0.9\ T$) then refining color patterns and textures ($0.8\ T$). 
Beyond $\tau=0.8\ T$, late-stage color fusion dominates, producing subjects almost completely identical to the reference subject ($0.7\ T$).
Quantitative results are shown in \cref{tab:ablation_study}, where the balanced $\tau=0.8\ T$ achieves 0.882 reference similarity preservation (CLIP-I) while maintaining 0.302 image-text alignment (CLIP-T).

\cparagraph{Effects of Patch Perturbation.}
By combining local token shuffling and mask morphing, our perturbation strategy reduces both texture and structure overfitting.
With $\tau=0.8\ T$, the generated subject and the reference one are structurally similar without perturbation (\cref{fig:ablation_study}).
Conversely, applying perturbation makes the structure and texture more flexible while maintaining identically consistency.

\subsection{Applications}

As illustrated in \cref{fig:method_extensions}, our \textit{Personalize Anything} naturally extends to diverse real-world applications, including subject-driven image generation with layout guidance, inpainting and outpainting.
Visualization results in \cref{fig:applications} demonstrate capabilities of our framework on layout-guided personalization, and precise editing with mask conditions, all without architectural modification or fine-tuning.

\section{Conclusion}
\label{sec:conclusion}

This paper reveals that simple token replacement achieves high-fidelity subject reconstruction in diffusion transformers (DiTs), due to the position-disentangled representation in DiTs.
The decoupling of semantic features and position enables the substitution of purely semantic tokens, avoiding positional interference.
Based on this discovery, we propose \textit{Personalize Anything}, a training-free framework that achieves high-fidelity personalized image generation through timestep-adaptive token injection and strategic patch perturbation.
Our method eliminates per-subject optimization or large-scale training while delivering superior identity preservation and unprecedented scalability to layout-guided generation, multi-subject personalization and mask-controlled editing.
DiTs' geometric programming establishes new paradigms for controllable synthesis, with spatial manipulation principles extensible to video/3D generation, redefining scalable customization in generative AI.

{
    \small
    \bibliographystyle{ieeenat_fullname}
    \bibliography{main}

\begin{thebibliography}{72}
\providecommand{\natexlab}[1]{#1}
\providecommand{\url}[1]{\texttt{#1}}
\expandafter\ifx\csname urlstyle\endcsname\relax
  \providecommand{\doi}[1]{doi: #1}\else
  \providecommand{\doi}{doi: \begingroup \urlstyle{rm}\Url}\fi

\bibitem[Avrahami et~al.(2024)Avrahami, Hertz, Vinker, Arar, Fruchter, Fried, Cohen-Or, and Lischinski]{ChosenOne}
Omri Avrahami, Amir Hertz, Yael Vinker, Moab Arar, Shlomi Fruchter, Ohad Fried, Daniel Cohen-Or, and Dani Lischinski.
\newblock The chosen one: Consistent characters in text-to-image diffusion models.
\newblock In \emph{Special Interest Group on Computer Graphics and Interactive Techniques Conference Conference Papers ’24}, page 1–12. ACM, 2024.

\bibitem[Chen et~al.(2024{\natexlab{a}})Chen, Huang, Liu, Shen, Zhao, and Zhao]{chen2024anydoor}
Xi Chen, Lianghua Huang, Yu Liu, Yujun Shen, Deli Zhao, and Hengshuang Zhao.
\newblock Anydoor: Zero-shot object-level image customization.
\newblock In \emph{Proceedings of the IEEE/CVF Conference on Computer Vision and Pattern Recognition}, pages 6593--6602, 2024{\natexlab{a}}.

\bibitem[Chen et~al.(2024{\natexlab{b}})Chen, Zhang, Zhang, Zhou, Kim, Liu, Li, Zhang, Zhao, Wang, et~al.]{UniReal}
Xi Chen, Zhifei Zhang, He Zhang, Yuqian Zhou, Soo~Ye Kim, Qing Liu, Yijun Li, Jianming Zhang, Nanxuan Zhao, Yilin Wang, et~al.
\newblock Unireal: Universal image generation and editing via learning real-world dynamics.
\newblock \emph{arXiv preprint arXiv:2412.07774}, 2024{\natexlab{b}}.

\bibitem[Dhariwal and Nichol(2021)]{dhariwal2021diffusionbeatgans}
Prafulla Dhariwal and Alexander Nichol.
\newblock Diffusion models beat gans on image synthesis.
\newblock \emph{Advances in neural information processing systems}, 34:\penalty0 8780--8794, 2021.

\bibitem[Ding et~al.(2024)Ding, Zhao, Wang, Yang, Liu, Chen, and Shen]{FreeCustom}
Ganggui Ding, Canyu Zhao, Wen Wang, Zhen Yang, Zide Liu, Hao Chen, and Chunhua Shen.
\newblock Freecustom: Tuning-free customized image generation for multi-concept composition.
\newblock In \emph{Proceedings of the IEEE/CVF Conference on Computer Vision and Pattern Recognition}, pages 9089--9098, 2024.

\bibitem[Dosovitskiy(2020)]{dosovitskiy2020vit}
Alexey Dosovitskiy.
\newblock An image is worth 16x16 words: Transformers for image recognition at scale.
\newblock \emph{arXiv preprint arXiv:2010.11929}, 2020.

\bibitem[Duan et~al.(2024)Duan, Ding, Gou, Zhou, Smith, and Liu]{EZIGen}
Zicheng Duan, Yuxuan Ding, Chenhui Gou, Ziqin Zhou, Ethan Smith, and Lingqiao Liu.
\newblock Ezigen: Enhancing zero-shot subject-driven image generation with precise subject encoding and decoupled guidance.
\newblock \emph{arXiv preprint arXiv:2409.08091}, 2024.

\bibitem[Esser et~al.(2024)Esser, Kulal, Blattmann, Entezari, Müller, Saini, Levi, Lorenz, Sauer, Boesel, Podell, Dockhorn, English, Lacey, Goodwin, Marek, and Rombach]{sd3}
Patrick Esser, Sumith Kulal, Andreas Blattmann, Rahim Entezari, Jonas Müller, Harry Saini, Yam Levi, Dominik Lorenz, Axel Sauer, Frederic Boesel, Dustin Podell, Tim Dockhorn, Zion English, Kyle Lacey, Alex Goodwin, Yannik Marek, and Robin Rombach.
\newblock Scaling rectified flow transformers for high-resolution image synthesis, 2024.

\bibitem[Fu et~al.(2023)Fu, Tamir, Sundaram, Chai, Zhang, Dekel, and Isola]{dreamsim}
Stephanie Fu, Netanel Tamir, Shobhita Sundaram, Lucy Chai, Richard Zhang, Tali Dekel, and Phillip Isola.
\newblock Dreamsim: Learning new dimensions of human visual similarity using synthetic data, 2023.

\bibitem[Gal et~al.(2022)Gal, Alaluf, Atzmon, Patashnik, Bermano, Chechik, and Cohen-Or]{oneimgoneword}
Rinon Gal, Yuval Alaluf, Yuval Atzmon, Or Patashnik, Amit~H. Bermano, Gal Chechik, and Daniel Cohen-Or.
\newblock An image is worth one word: Personalizing text-to-image generation using textual inversion, 2022.

\bibitem[Gal et~al.(2024)Gal, Lichter, Richardson, Patashnik, Bermano, Chechik, and Cohen-Or]{LcmLookahead}
Rinon Gal, Or Lichter, Elad Richardson, Or Patashnik, Amit~H Bermano, Gal Chechik, and Daniel Cohen-Or.
\newblock Lcm-lookahead for encoder-based text-to-image personalization.
\newblock In \emph{European Conference on Computer Vision}, pages 322--340. Springer, 2024.

\bibitem[Gu et~al.(2024)Gu, Wang, Wu, Shi, Chen, Fan, Xiao, Zhao, Chang, Wu, et~al.]{MixOfShow}
Yuchao Gu, Xintao Wang, Jay~Zhangjie Wu, Yujun Shi, Yunpeng Chen, Zihan Fan, Wuyou Xiao, Rui Zhao, Shuning Chang, Weijia Wu, et~al.
\newblock Mix-of-show: Decentralized low-rank adaptation for multi-concept customization of diffusion models.
\newblock \emph{Advances in Neural Information Processing Systems}, 36, 2024.

\bibitem[Guo et~al.(2024)Guo, Wu, Chen, Chen, Zhang, and He]{Pulid}
Zinan Guo, Yanze Wu, Zhuowei Chen, Lang Chen, Peng Zhang, and Qian He.
\newblock Pulid: Pure and lightning id customization via contrastive alignment.
\newblock \emph{arXiv preprint arXiv:2404.16022}, 2024.

\bibitem[Han et~al.(2023)Han, Li, Zhang, Milanfar, Metaxas, and Yang]{svdiff}
Ligong Han, Yinxiao Li, Han Zhang, Peyman Milanfar, Dimitris Metaxas, and Feng Yang.
\newblock Svdiff: Compact parameter space for diffusion fine-tuning, 2023.

\bibitem[He et~al.(2025)He, Tuo, Chen, Zhong, Geng, and Bo]{AnyStory}
Junjie He, Yuxiang Tuo, Binghui Chen, Chongyang Zhong, Yifeng Geng, and Liefeng Bo.
\newblock Anystory: Towards unified single and multiple subject personalization in text-to-image generation.
\newblock \emph{arXiv preprint arXiv:2501.09503}, 2025.

\bibitem[Heusel et~al.(2018)Heusel, Ramsauer, Unterthiner, Nessler, and Hochreiter]{fid}
Martin Heusel, Hubert Ramsauer, Thomas Unterthiner, Bernhard Nessler, and Sepp Hochreiter.
\newblock Gans trained by a two time-scale update rule converge to a local nash equilibrium, 2018.

\bibitem[Ho et~al.(2020)Ho, Jain, and Abbeel]{ddpm}
Jonathan Ho, Ajay Jain, and Pieter Abbeel.
\newblock Denoising diffusion probabilistic models, 2020.

\bibitem[Huang et~al.(2024{\natexlab{a}})Huang, Dong, Song, Li, Zhou, Cheng, Liao, Chen, Yan, Liao, et~al.]{Consistentid}
Jiehui Huang, Xiao Dong, Wenhui Song, Hanhui Li, Jun Zhou, Yuhao Cheng, Shutao Liao, Long Chen, Yiqiang Yan, Shengcai Liao, et~al.
\newblock Consistentid: Portrait generation with multimodal fine-grained identity preserving.
\newblock \emph{arXiv preprint arXiv:2404.16771}, 2024{\natexlab{a}}.

\bibitem[Huang et~al.(2024{\natexlab{b}})Huang, Fu, Liu, Jiang, Yu, and Song]{mipadapter}
Qihan Huang, Siming Fu, Jinlong Liu, Hao Jiang, Yipeng Yu, and Jie Song.
\newblock Resolving multi-condition confusion for finetuning-free personalized image generation.
\newblock \emph{arXiv preprint arXiv:2409.17920}, 2024{\natexlab{b}}.

\bibitem[Huang et~al.(2024{\natexlab{c}})Huang, Fan, Wang, and Sheng]{Parts2whole}
Zehuan Huang, Hongxing Fan, Lipeng Wang, and Lu Sheng.
\newblock From parts to whole: A unified reference framework for controllable human image generation, 2024{\natexlab{c}}.

\bibitem[Hyung et~al.(2024)Hyung, Shin, and Choo]{Magicapture}
Junha Hyung, Jaeyo Shin, and Jaegul Choo.
\newblock Magicapture: High-resolution multi-concept portrait customization.
\newblock In \emph{Proceedings of the AAAI Conference on Artificial Intelligence}, pages 2445--2453, 2024.

\bibitem[Jiang et~al.(2024)Jiang, Zhang, Feng, Wu, and Zuo]{mc2}
Jiaxiu Jiang, Yabo Zhang, Kailai Feng, Xiaohe Wu, and Wangmeng Zuo.
\newblock Mc$^2$: Multi-concept guidance for customized multi-concept generation.
\newblock \emph{arXiv preprint arXiv:2404.05268}, 2024.

\bibitem[Kong et~al.(2024)Kong, Zhang, Yang, Wang, Zhang, Wu, Chen, Liu, and Luo]{Omg}
Zhe Kong, Yong Zhang, Tianyu Yang, Tao Wang, Kaihao Zhang, Bizhu Wu, Guanying Chen, Wei Liu, and Wenhan Luo.
\newblock Omg: Occlusion-friendly personalized multi-concept generation in diffusion models.
\newblock In \emph{European Conference on Computer Vision}, pages 253--270. Springer, 2024.

\bibitem[Kumari et~al.(2023)Kumari, Zhang, Zhang, Shechtman, and Zhu]{MultiConcept}
Nupur Kumari, Bingliang Zhang, Richard Zhang, Eli Shechtman, and Jun-Yan Zhu.
\newblock Multi-concept customization of text-to-image diffusion.
\newblock In \emph{Proceedings of the IEEE/CVF Conference on Computer Vision and Pattern Recognition}, pages 1931--1941, 2023.

\bibitem[Kwon and Ye(2025)]{TweedieMix}
Gihyun Kwon and Jong~Chul Ye.
\newblock Tweediemix: Improving multi-concept fusion for diffusion-based image/video generation, 2025.

\bibitem[Kwon et~al.(2024)Kwon, Jenni, Li, Lee, Ye, and Heilbron]{ConceptWeaver}
Gihyun Kwon, Simon Jenni, Dingzeyu Li, Joon-Young Lee, Jong~Chul Ye, and Fabian~Caba Heilbron.
\newblock Concept weaver: Enabling multi-concept fusion in text-to-image models.
\newblock In \emph{Proceedings of the IEEE/CVF Conference on Computer Vision and Pattern Recognition}, pages 8880--8889, 2024.

\bibitem[Labs(2024)]{flux}
Black~Forest Labs.
\newblock Flux.
\newblock [Online], 2024.
\newblock \url{https://github.com/black-forest-labs/flux}.

\bibitem[Le et~al.(2024)Le, Pham, Lee, Clark, Kembhavi, Mandt, Krishna, and Lu]{OneDiffusion}
Duong~H. Le, Tuan Pham, Sangho Lee, Christopher Clark, Aniruddha Kembhavi, Stephan Mandt, Ranjay Krishna, and Jiasen Lu.
\newblock One diffusion to generate them all, 2024.

\bibitem[Li et~al.(2024{\natexlab{a}})Li, Li, and Hoi]{BlipDiffusion}
Dongxu Li, Junnan Li, and Steven Hoi.
\newblock Blip-diffusion: Pre-trained subject representation for controllable text-to-image generation and editing.
\newblock \emph{Advances in Neural Information Processing Systems}, 36, 2024{\natexlab{a}}.

\bibitem[Li et~al.(2024{\natexlab{b}})Li, Cao, Wang, Qi, Cheng, and Shan]{Photomaker}
Zhen Li, Mingdeng Cao, Xintao Wang, Zhongang Qi, Ming-Ming Cheng, and Ying Shan.
\newblock Photomaker: Customizing realistic human photos via stacked id embedding.
\newblock In \emph{Proceedings of the IEEE/CVF Conference on Computer Vision and Pattern Recognition}, pages 8640--8650, 2024{\natexlab{b}}.

\bibitem[Li et~al.(2024{\natexlab{c}})Li, Zhang, Lin, Xiong, Long, Deng, Zhang, Liu, Huang, Xiao, Chen, He, Li, Li, Zhang, Quan, Lu, Huang, Yuan, Zheng, Li, Zhang, Zhang, Chen, Liu, Fang, Wang, Xue, Tao, Zhu, Liu, Lin, Sun, Li, Wang, Chen, Hu, Xiao, Chen, Liu, Liu, Wang, Yang, Jiang, and Lu]{hunyuandit}
Zhimin Li, Jianwei Zhang, Qin Lin, Jiangfeng Xiong, Yanxin Long, Xinchi Deng, Yingfang Zhang, Xingchao Liu, Minbin Huang, Zedong Xiao, Dayou Chen, Jiajun He, Jiahao Li, Wenyue Li, Chen Zhang, Rongwei Quan, Jianxiang Lu, Jiabin Huang, Xiaoyan Yuan, Xiaoxiao Zheng, Yixuan Li, Jihong Zhang, Chao Zhang, Meng Chen, Jie Liu, Zheng Fang, Weiyan Wang, Jinbao Xue, Yangyu Tao, Jianchen Zhu, Kai Liu, Sihuan Lin, Yifu Sun, Yun Li, Dongdong Wang, Mingtao Chen, Zhichao Hu, Xiao Xiao, Yan Chen, Yuhong Liu, Wei Liu, Di Wang, Yong Yang, Jie Jiang, and Qinglin Lu.
\newblock Hunyuan-dit: A powerful multi-resolution diffusion transformer with fine-grained chinese understanding, 2024{\natexlab{c}}.

\bibitem[Liu et~al.(2023)Liu, Zhang, Shen, Zheng, Zhu, Feng, Liu, Zhao, Zhou, and Cao]{Cones2}
Zhiheng Liu, Yifei Zhang, Yujun Shen, Kecheng Zheng, Kai Zhu, Ruili Feng, Yu Liu, Deli Zhao, Jingren Zhou, and Yang Cao.
\newblock Cones 2: Customizable image synthesis with multiple subjects.
\newblock In \emph{Proceedings of the 37th International Conference on Neural Information Processing Systems}, pages 57500--57519, 2023.

\bibitem[Ma et~al.(2024)Ma, Liang, Chen, and Lu]{SubjectDiffusion}
Jian Ma, Junhao Liang, Chen Chen, and Haonan Lu.
\newblock Subject-diffusion: Open domain personalized text-to-image generation without test-time fine-tuning.
\newblock In \emph{ACM SIGGRAPH 2024 Conference Papers}, pages 1--12, 2024.

\bibitem[Nichol et~al.(2022)Nichol, Dhariwal, Ramesh, Shyam, Mishkin, McGrew, Sutskever, and Chen]{nichol2022glide}
Alex Nichol, Prafulla Dhariwal, Aditya Ramesh, Pranav Shyam, Pamela Mishkin, Bob McGrew, Ilya Sutskever, and Mark Chen.
\newblock Glide: Towards photorealistic image generation and editing with text-guided diffusion models, 2022.

\bibitem[Oquab et~al.(2024)Oquab, Darcet, Moutakanni, Vo, Szafraniec, Khalidov, Fernandez, Haziza, Massa, El-Nouby, Assran, Ballas, Galuba, Howes, Huang, Li, Misra, Rabbat, Sharma, Synnaeve, Xu, Jegou, Mairal, Labatut, Joulin, and Bojanowski]{dinov2}
Maxime Oquab, Timothée Darcet, Théo Moutakanni, Huy Vo, Marc Szafraniec, Vasil Khalidov, Pierre Fernandez, Daniel Haziza, Francisco Massa, Alaaeldin El-Nouby, Mahmoud Assran, Nicolas Ballas, Wojciech Galuba, Russell Howes, Po-Yao Huang, Shang-Wen Li, Ishan Misra, Michael Rabbat, Vasu Sharma, Gabriel Synnaeve, Hu Xu, Hervé Jegou, Julien Mairal, Patrick Labatut, Armand Joulin, and Piotr Bojanowski.
\newblock Dinov2: Learning robust visual features without supervision, 2024.

\bibitem[Parmar et~al.(2025)Parmar, Patashnik, Wang, Ostashev, Narasimhan, Zhu, Cohen-Or, and Aberman]{ObjectLevel}
Gaurav Parmar, Or Patashnik, Kuan-Chieh Wang, Daniil Ostashev, Srinivasa Narasimhan, Jun-Yan Zhu, Daniel Cohen-Or, and Kfir Aberman.
\newblock Object-level visual prompts for compositional image generation.
\newblock \emph{arXiv preprint arXiv:2501.01424}, 2025.

\bibitem[Patashnik et~al.(2025)Patashnik, Gal, Ostashev, Tulyakov, Aberman, and Cohen-Or]{NestedAttention}
Or Patashnik, Rinon Gal, Daniil Ostashev, Sergey Tulyakov, Kfir Aberman, and Daniel Cohen-Or.
\newblock Nested attention: Semantic-aware attention values for concept personalization.
\newblock \emph{arXiv preprint arXiv:2501.01407}, 2025.

\bibitem[Patel et~al.(2024)Patel, Jung, Baral, and Yang]{lambdaECLIPSE}
Maitreya Patel, Sangmin Jung, Chitta Baral, and Yezhou Yang.
\newblock $\lambda$-eclipse: Multi-concept personalized text-to-image diffusion models by leveraging clip latent space, 2024.

\bibitem[Peebles and Xie(2023)]{peebles2023dit}
William Peebles and Saining Xie.
\newblock Scalable diffusion models with transformers.
\newblock In \emph{Proceedings of the IEEE/CVF International Conference on Computer Vision}, pages 4195--4205, 2023.

\bibitem[Podell et~al.(2023)Podell, English, Lacey, Blattmann, Dockhorn, M{\"u}ller, Penna, and Rombach]{podell2023sdxl}
Dustin Podell, Zion English, Kyle Lacey, Andreas Blattmann, Tim Dockhorn, Jonas M{\"u}ller, Joe Penna, and Robin Rombach.
\newblock Sdxl: Improving latent diffusion models for high-resolution image synthesis.
\newblock \emph{arXiv preprint arXiv:2307.01952}, 2023.

\bibitem[Radford et~al.(2021)Radford, Kim, Hallacy, Ramesh, Goh, Agarwal, Sastry, Askell, Mishkin, Clark, Krueger, and Sutskever]{clip}
Alec Radford, Jong~Wook Kim, Chris Hallacy, Aditya Ramesh, Gabriel Goh, Sandhini Agarwal, Girish Sastry, Amanda Askell, Pamela Mishkin, Jack Clark, Gretchen Krueger, and Ilya Sutskever.
\newblock Learning transferable visual models from natural language supervision, 2021.

\bibitem[Ramesh et~al.(2022)Ramesh, Dhariwal, Nichol, Chu, and Chen]{ramesh2022dalle2}
Aditya Ramesh, Prafulla Dhariwal, Alex Nichol, Casey Chu, and Mark Chen.
\newblock Hierarchical text-conditional image generation with clip latents.
\newblock \emph{arXiv preprint arXiv:2204.06125}, 1\penalty0 (2):\penalty0 3, 2022.

\bibitem[Richardson et~al.(2024)Richardson, Alaluf, Mahdavi-Amiri, and Cohen-Or]{pOps}
Elad Richardson, Yuval Alaluf, Ali Mahdavi-Amiri, and Daniel Cohen-Or.
\newblock pops: Photo-inspired diffusion operators.
\newblock \emph{arXiv preprint arXiv:2406.01300}, 2024.

\bibitem[Rombach et~al.(2022)Rombach, Blattmann, Lorenz, Esser, and Ommer]{ldm}
Robin Rombach, Andreas Blattmann, Dominik Lorenz, Patrick Esser, and Björn Ommer.
\newblock High-resolution image synthesis with latent diffusion models, 2022.

\bibitem[Ronneberger et~al.(2015)Ronneberger, Fischer, and Brox]{unet}
Olaf Ronneberger, Philipp Fischer, and Thomas Brox.
\newblock U-net: Convolutional networks for biomedical image segmentation, 2015.

\bibitem[Rout et~al.(2024)Rout, Chen, Ruiz, Caramanis, Shakkottai, and Chu]{SemanticImageInversion}
Litu Rout, Yujia Chen, Nataniel Ruiz, Constantine Caramanis, Sanjay Shakkottai, and Wen-Sheng Chu.
\newblock Semantic image inversion and editing using rectified stochastic differential equations.
\newblock \emph{arXiv preprint arXiv:2410.10792}, 2024.

\bibitem[Ruiz et~al.(2023)Ruiz, Li, Jampani, Pritch, Rubinstein, and Aberman]{Dreambooth}
Nataniel Ruiz, Yuanzhen Li, Varun Jampani, Yael Pritch, Michael Rubinstein, and Kfir Aberman.
\newblock Dreambooth: Fine tuning text-to-image diffusion models for subject-driven generation.
\newblock In \emph{Proceedings of the IEEE/CVF conference on computer vision and pattern recognition}, pages 22500--22510, 2023.

\bibitem[Saharia et~al.(2022)Saharia, Chan, Saxena, Li, Whang, Denton, Ghasemipour, Gontijo~Lopes, Karagol~Ayan, Salimans, et~al.]{saharia2022imagen}
Chitwan Saharia, William Chan, Saurabh Saxena, Lala Li, Jay Whang, Emily~L Denton, Kamyar Ghasemipour, Raphael Gontijo~Lopes, Burcu Karagol~Ayan, Tim Salimans, et~al.
\newblock Photorealistic text-to-image diffusion models with deep language understanding.
\newblock \emph{Advances in neural information processing systems}, 35:\penalty0 36479--36494, 2022.

\bibitem[Shah et~al.(2024)Shah, Ruiz, Cole, Lu, Lazebnik, Li, and Jampani]{Ziplora}
Viraj Shah, Nataniel Ruiz, Forrester Cole, Erika Lu, Svetlana Lazebnik, Yuanzhen Li, and Varun Jampani.
\newblock Ziplora: Any subject in any style by effectively merging loras.
\newblock In \emph{European Conference on Computer Vision}, pages 422--438. Springer, 2024.

\bibitem[Shin et~al.(2024)Shin, Choi, Kim, and Yoon]{LargeScale}
Chaehun Shin, Jooyoung Choi, Heeseung Kim, and Sungroh Yoon.
\newblock Large-scale text-to-image model with inpainting is a zero-shot subject-driven image generator.
\newblock \emph{arXiv preprint arXiv:2411.15466}, 2024.

\bibitem[Song et~al.(2022)Song, Meng, and Ermon]{ddim}
Jiaming Song, Chenlin Meng, and Stefano Ermon.
\newblock Denoising diffusion implicit models, 2022.

\bibitem[Song et~al.(2024)Song, Zhu, Liu, Yan, Elgammal, and Yang]{Moma}
Kunpeng Song, Yizhe Zhu, Bingchen Liu, Qing Yan, Ahmed Elgammal, and Xiao Yang.
\newblock Moma: Multimodal llm adapter for fast personalized image generation.
\newblock In \emph{European Conference on Computer Vision}, pages 117--132. Springer, 2024.

\bibitem[Su et~al.(2024)Su, Ahmed, Lu, Pan, Bo, and Liu]{su2024rope}
Jianlin Su, Murtadha Ahmed, Yu Lu, Shengfeng Pan, Wen Bo, and Yunfeng Liu.
\newblock Roformer: Enhanced transformer with rotary position embedding.
\newblock \emph{Neurocomputing}, 568:\penalty0 127063, 2024.

\bibitem[Tan et~al.(2024)Tan, Liu, Yang, Xue, and Wang]{Ominicontrol}
Zhenxiong Tan, Songhua Liu, Xingyi Yang, Qiaochu Xue, and Xinchao Wang.
\newblock Ominicontrol: Minimal and universal control for diffusion transformer.
\newblock \emph{arXiv preprint arXiv:2411.15098}, 3, 2024.

\bibitem[Team(2024)]{kolors}
Kolors Team.
\newblock Kolors: Effective training of diffusion model for photorealistic text-to-image synthesis.
\newblock \emph{arXiv preprint}, 2024.

\bibitem[Tewel et~al.(2024{\natexlab{a}})Tewel, Gal, Chechik, and Atzmon]{keylocked}
Yoad Tewel, Rinon Gal, Gal Chechik, and Yuval Atzmon.
\newblock Key-locked rank one editing for text-to-image personalization, 2024{\natexlab{a}}.

\bibitem[Tewel et~al.(2024{\natexlab{b}})Tewel, Kaduri, Gal, Kasten, Wolf, Chechik, and Atzmon]{consistory}
Yoad Tewel, Omri Kaduri, Rinon Gal, Yoni Kasten, Lior Wolf, Gal Chechik, and Yuval Atzmon.
\newblock Training-free consistent text-to-image generation.
\newblock \emph{ACM Transactions on Graphics (TOG)}, 43\penalty0 (4):\penalty0 1--18, 2024{\natexlab{b}}.

\bibitem[Voynov et~al.(2023)Voynov, Chu, Cohen-Or, and Aberman]{pplus}
Andrey Voynov, Qinghao Chu, Daniel Cohen-Or, and Kfir Aberman.
\newblock P+: Extended textual conditioning in text-to-image generation, 2023.

\bibitem[Wang et~al.(2023)Wang, Li, Zhang, Xu, Zhou, Yu, Sheng, and Xu]{DiffusionIsSegmenter}
Jinglong Wang, Xiawei Li, Jing Zhang, Qingyuan Xu, Qin Zhou, Qian Yu, Lu Sheng, and Dong Xu.
\newblock Diffusion model is secretly a training-free open vocabulary semantic segmenter.
\newblock \emph{arXiv preprint arXiv:2309.02773}, 2023.

\bibitem[Wang et~al.(2024{\natexlab{a}})Wang, Pu, Qi, Guo, Ma, Huang, Chen, Li, and Shan]{TamingRectifiedFlow}
Jiangshan Wang, Junfu Pu, Zhongang Qi, Jiayi Guo, Yue Ma, Nisha Huang, Yuxin Chen, Xiu Li, and Ying Shan.
\newblock Taming rectified flow for inversion and editing.
\newblock \emph{arXiv preprint arXiv:2411.04746}, 2024{\natexlab{a}}.

\bibitem[Wang et~al.(2024{\natexlab{b}})Wang, Bai, Wang, Qin, Chen, Li, Tang, and Hu]{Instantid}
Qixun Wang, Xu Bai, Haofan Wang, Zekui Qin, Anthony Chen, Huaxia Li, Xu Tang, and Yao Hu.
\newblock Instantid: Zero-shot identity-preserving generation in seconds.
\newblock \emph{arXiv preprint arXiv:2401.07519}, 2024{\natexlab{b}}.

\bibitem[Wang et~al.(2024{\natexlab{c}})Wang, Li, Li, Cao, Ma, Lu, and Jia]{CharacterFactory}
Qinghe Wang, Baolu Li, Xiaomin Li, Bing Cao, Liqian Ma, Huchuan Lu, and Xu Jia.
\newblock Characterfactory: Sampling consistent characters with gans for diffusion models.
\newblock \emph{arXiv preprint arXiv:2404.15677}, 2024{\natexlab{c}}.

\bibitem[Wang et~al.(2025)Wang, Fu, Huang, He, and Jiang]{msdiffusion}
X. Wang, Siming Fu, Qihan Huang, Wanggui He, and Hao Jiang.
\newblock Ms-diffusion: Multi-subject zero-shot image personalization with layout guidance, 2025.

\bibitem[Xing et~al.(2024)Xing, Wang, Sun, Wang, Bai, Ai, Huang, and Li]{CSGO}
Peng Xing, Haofan Wang, Yanpeng Sun, Qixun Wang, Xu Bai, Hao Ai, Renyuan Huang, and Zechao Li.
\newblock Csgo: Content-style composition in text-to-image generation, 2024.

\bibitem[Ye et~al.(2023)Ye, Zhang, Liu, Han, and Yang]{IpAdapter}
Hu Ye, Jun Zhang, Sibo Liu, Xiao Han, and Wei Yang.
\newblock Ip-adapter: Text compatible image prompt adapter for text-to-image diffusion models.
\newblock \emph{arXiv preprint arXiv:2308.06721}, 2023.

\bibitem[Zhang et~al.(2018)Zhang, Isola, Efros, Shechtman, and Wang]{lpips}
Richard Zhang, Phillip Isola, Alexei~A. Efros, Eli Shechtman, and Oliver Wang.
\newblock The unreasonable effectiveness of deep features as a perceptual metric, 2018.

\bibitem[Zhang et~al.(2024{\natexlab{a}})Zhang, Huang, Chen, Zhang, Wu, Feng, Wang, Shen, Liu, and Luo]{FlashFace}
Shilong Zhang, Lianghua Huang, Xi Chen, Yifei Zhang, Zhi-Fan Wu, Yutong Feng, Wei Wang, Yujun Shen, Yu Liu, and Ping Luo.
\newblock Flashface: Human image personalization with high-fidelity identity preservation.
\newblock \emph{arXiv preprint arXiv:2403.17008}, 2024{\natexlab{a}}.

\bibitem[Zhang et~al.(2024{\natexlab{b}})Zhang, Song, Liu, Wang, Yu, Tang, Li, Tang, Hu, Pan, et~al.]{SsrEncoder}
Yuxuan Zhang, Yiren Song, Jiaming Liu, Rui Wang, Jinpeng Yu, Hao Tang, Huaxia Li, Xu Tang, Yao Hu, Han Pan, et~al.
\newblock Ssr-encoder: Encoding selective subject representation for subject-driven generation.
\newblock In \emph{Proceedings of the IEEE/CVF Conference on Computer Vision and Pattern Recognition}, pages 8069--8078, 2024{\natexlab{b}}.

\bibitem[Zhang et~al.(2025)Zhang, Luo, Dong, Yang, Huang, Ma, Deussen, Lee, and Xu]{BC2NewStories}
Yuxin Zhang, Minyan Luo, Weiming Dong, Xiao Yang, Haibin Huang, Chongyang Ma, Oliver Deussen, Tong-Yee Lee, and Changsheng Xu.
\newblock Bringing characters to new stories: Training-free theme-specific image generation via dynamic visual prompting.
\newblock \emph{arXiv preprint arXiv:2501.15641}, 2025.

\bibitem[Zhou et~al.(2024)Zhou, Zhou, Cheng, Feng, and Hou]{StoryDiffusion}
Yupeng Zhou, Daquan Zhou, Ming-Ming Cheng, Jiashi Feng, and Qibin Hou.
\newblock Storydiffusion: Consistent self-attention for long-range image and video generation.
\newblock \emph{arXiv preprint arXiv:2405.01434}, 2024.

\bibitem[Zhu et~al.(2024)Zhu, Li, Ma, He, and Xiu]{multibooth}
Chenyang Zhu, Kai Li, Yue Ma, Chunming He, and Li Xiu.
\newblock Multibooth: Towards generating all your concepts in an image from text.
\newblock \emph{arXiv preprint arXiv:2404.14239}, 2024.

\bibitem[Zong et~al.(2024)Zong, Jiang, Ma, Song, Shao, Shen, Liu, and Li]{Easyref}
Zhuofan Zong, Dongzhi Jiang, Bingqi Ma, Guanglu Song, Hao Shao, Dazhong Shen, Yu Liu, and Hongsheng Li.
\newblock Easyref: Omni-generalized group image reference for diffusion models via multimodal llm.
\newblock \emph{arXiv preprint arXiv:2412.09618}, 2024.

\end{thebibliography}
}

\clearpage
\setcounter{page}{1}
\maketitlesupplementary

\section{Analysis of Position Sensitivity in DiT}
\label{sec:analysis_position}
We employ the attention sharing mechanism in existing UNet-based subject personalization methods~\cite{consistory,FreeCustom} to diffusion transformers (DiTs)~\cite{peebles2023dit}.
Specifically, we concatenate position-encoded denoising tokens $\dot{X}$ and reference tokens $\dot{X}_{ref}$ into a single sequence $[\dot{X};\dot{X}{ref}]$ and apply pre-trained multi-modal attention on it.
In this process, both denoising and reference tokens retain their original positions $(i,j)\in [0,w)\times[0,h)$, causing destructive interference to attention computation and producing ghosting artifacts of the reference subject on the generated image.
For quantitative analysis, we calculate the denoising tokens' attention scores on the reference tokens at the same positions.
By averaging these scores across 100 samples, we obtain a mean value of 0.4294.
Subsequently, we apply the same procedure to U-Net, which encodes positional information by convolution layers instead of explicit positional encoding, yielding an average attention score of 0.0522.
Comparatively, the average attention score in DiT is \textbf{723\%} higher than that in U-Net.
The above quantitative analysis demonstrates the position sensitivity in DiT, where explicit positional encoding significantly influences the attention mechanism, contrasting with U-Net’s implicit position handling.

\section{User Study}
\label{sec:user_study}

\begin{figure}[h]
    \centering
    \includegraphics[width=\linewidth]{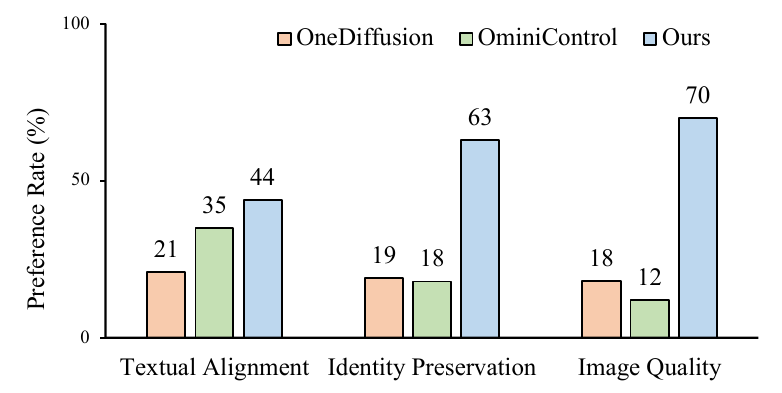}
    \caption{User study results on single-subject personalization.}
    \label{fig:user_study_single_subject}
\end{figure}

\begin{figure}[h]
    \centering
    \includegraphics[width=\linewidth]{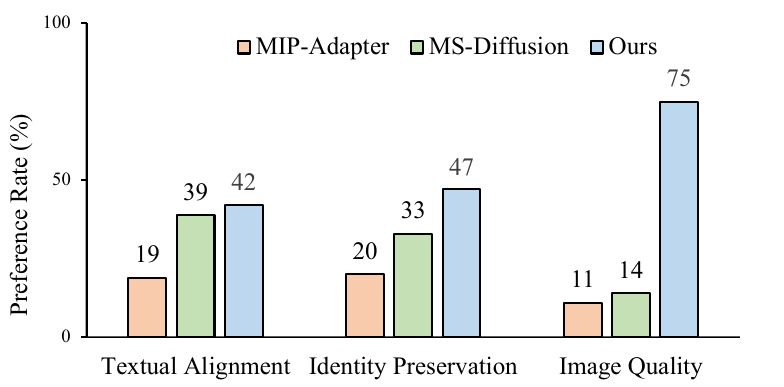}
    \caption{User study results on multi-subject personalization.}
    \label{fig:user_study_multi_subject}
\end{figure}

\begin{figure}[h]
    \centering
    \includegraphics[width=\linewidth]{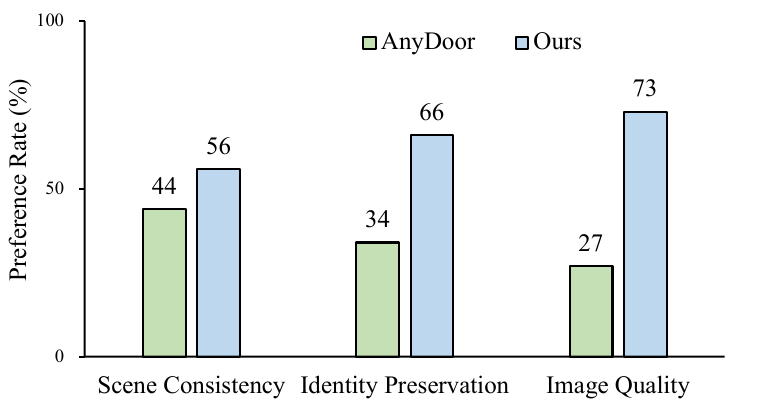}
    \caption{User study results on subject-scene composition.}
    \label{fig:user_study_subject_scene}
\end{figure}

To further evaluate model performance, we conducted a user study involving 48 participants, with ages evenly distributed between 15 and 60 years. 
Each participant was asked to answer 15 questions, resulting in a total of 720 valid responses.
For single-subject and multi-subject personalization tasks, users are required to select the optimal model based on three dimensions: textual alignment, identity preservation, and image quality. 
In subject-scene composition tasks, we substituted textual alignment with scene consistency to assess subject-scene coordination.
The results are presented in \cref{fig:user_study_single_subject,fig:user_study_multi_subject,fig:user_study_subject_scene}, which further corroborates our qualitative findings, as our method outperforms other state-of-the-art methods across all metrics.

\section{More Results}
\label{sec:more_results}

\begin{figure*}
    \centering
    \includegraphics[width=\textwidth]{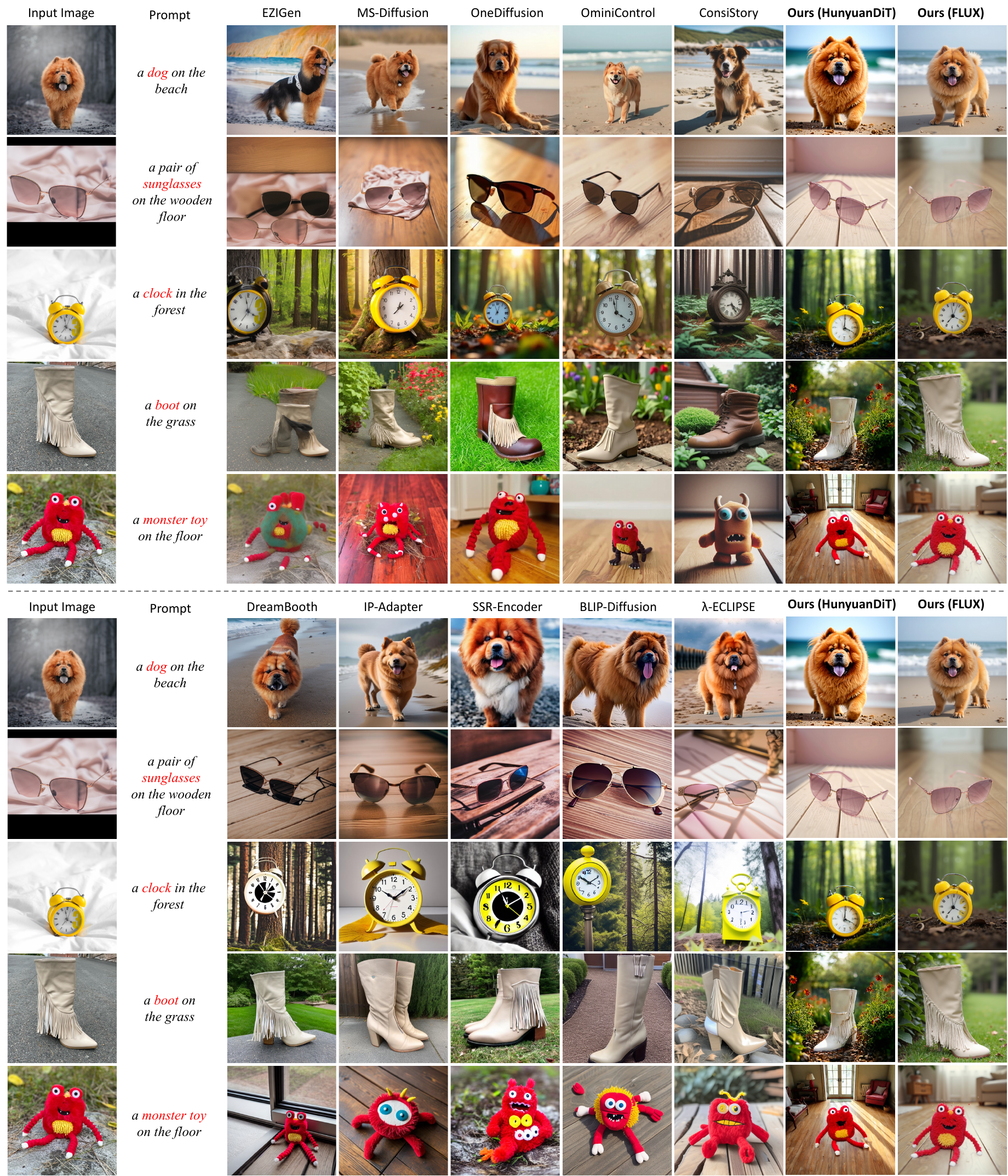}
    \caption{Full qualitative comparisons on single-subject personalization.}
    \label{fig:full_experiment}
\end{figure*}

We present more qualitative comparison results in \cref{fig:full_experiment}.
As demonstrated, our method consistently achieves exceptional textual alignment and subject consistency.
Specifically, our approach excels in preserving fine-grained details of subjects (\eg the number ``3'' on the clock) while maintaining high fidelity to the textual description. 

\end{document}